\documentclass[letterpaper, 10 pt, conference]{ieeeconf}  

\IEEEoverridecommandlockouts                              

\usepackage{algorithm}
\usepackage{algorithmic}

\usepackage{units}
\usepackage{subfigure}
\usepackage{url}
\usepackage{graphicx}
\usepackage{mathtools}
\usepackage{amsmath}
\usepackage{dsfont}
\usepackage{sidecap}
\usepackage{wrapfig}
\usepackage{amsmath,amssymb}
\usepackage{bm}
\usepackage{xcolor}
\usepackage{flushend}

\usepackage{color}

\newcommand{\twofig}{0.46\hsize}

\newcommand{\R}{\mathds{R}}
\newcommand{\Z}{\mathds{Z}}
\renewcommand{\vec}{\boldsymbol}
\newcommand{\mat}{\boldsymbol}
\newcommand{\E}{\mathds{E}}

\newcommand{\diag}{\mathrm{diag}}

\newcommand{\cost}{c}

\newcommand{\prob}{{p}}
\newcommand{\gauss}[2]{\mathcal N(#1,#2)}
\newcommand{\gaussx}[3]{\mathcal{N}\big(#1\,|\,#2,#3\big)}

\newcommand{\polpar}{\theta}
\newcommand{\target}{\eta}
\newcommand{\task}{\target}

\newcommand{\eq}{Eq.}

\newcommand{\fig}{Fig.}

\newcommand{\alg}{Alg.}

\newcommand{\pred}{^\pi}
\newcommand{\expert}{^{\text{exp}}}
\newcommand{\ffrac}[2]{{\frac{#1}{#2}}}

\definecolor{orange}{rgb}{1,0.5,0}
\definecolor{cyan}{rgb}{0,0.9,0.9}
\definecolor{magenta}{rgb}{1,0,1}
\definecolor{green2}{rgb}{0.2, 0.6, 0.2}

\newcommand{\red}[1]{\textcolor{red}{#1}}

\newcommand{\figspace}{\vspace{-4mm}}

\newcommand{\train}[0]{^{\text{train}}}
\newcommand{\test}[0]{^{\text{test}}}

\usepackage{graphicx} 
\usepackage{subfigure} 

\usepackage{algorithm}
\usepackage{algorithmic}

\title{\LARGE \bf
Multi-Task Policy Search
}

\author{Marc Peter Deisenroth$^{1,2}$, Peter Englert$^{3}$, Jan
  Peters$^{2,4}$, and Dieter Fox$^5$
  \thanks{A concise version of this paper has been published at the
    \emph{IEEE International Conference on Robotics and Automation (ICRA)
    2014~\cite{Deisenroth2014a}.}}  \thanks{The research leading to
    these results has received funding from the European Community's
    Seventh Framework Programme (FP7/2007-2013) under grant agreement
    \#270327, ONR MURI grant N00014-09-1-1052, and the Department of
    Computing, Imperial College London.}
  \thanks{$^{1}$Department of Computing, Imperial College London, UK}%
  \thanks{$^{2}$Department of Computer Science, TU Darmstadt,
    Germany}%
  \thanks{$^{3}$Department of Computer Science, University of
    Stuttgart, Germany}%
  \thanks{$^{4}$Max Planck Institute for Intelligent Systems,
    Germany}%
  \thanks{$^{5}$Department of Computer Science and Engineering,
    University of Washington, WA, USA}%
}

\begin{document}

\maketitle
\thispagestyle{empty}
\pagestyle{empty}


\begin{abstract}
  Learning policies that generalize across multiple tasks is an
  important and challenging research topic in reinforcement learning
  and robotics. Training individual policies for every single
  potential task is often impractical, especially for continuous
  task variations, requiring more principled approaches to share and
  transfer knowledge among similar tasks.  We present a novel approach
  for learning a nonlinear feedback policy that generalizes across
  multiple tasks. The key idea is to define a parametrized policy as a
  function of both the state \emph{and} the task, which allows
  learning a single policy that generalizes across multiple known and
  unknown tasks. Applications of our novel approach to reinforcement
  and imitation learning in real-robot experiments are shown.
\end{abstract}



\section{Introduction}

\begin{figure}[tb]
\centering
\includegraphics[width = 0.8\hsize]{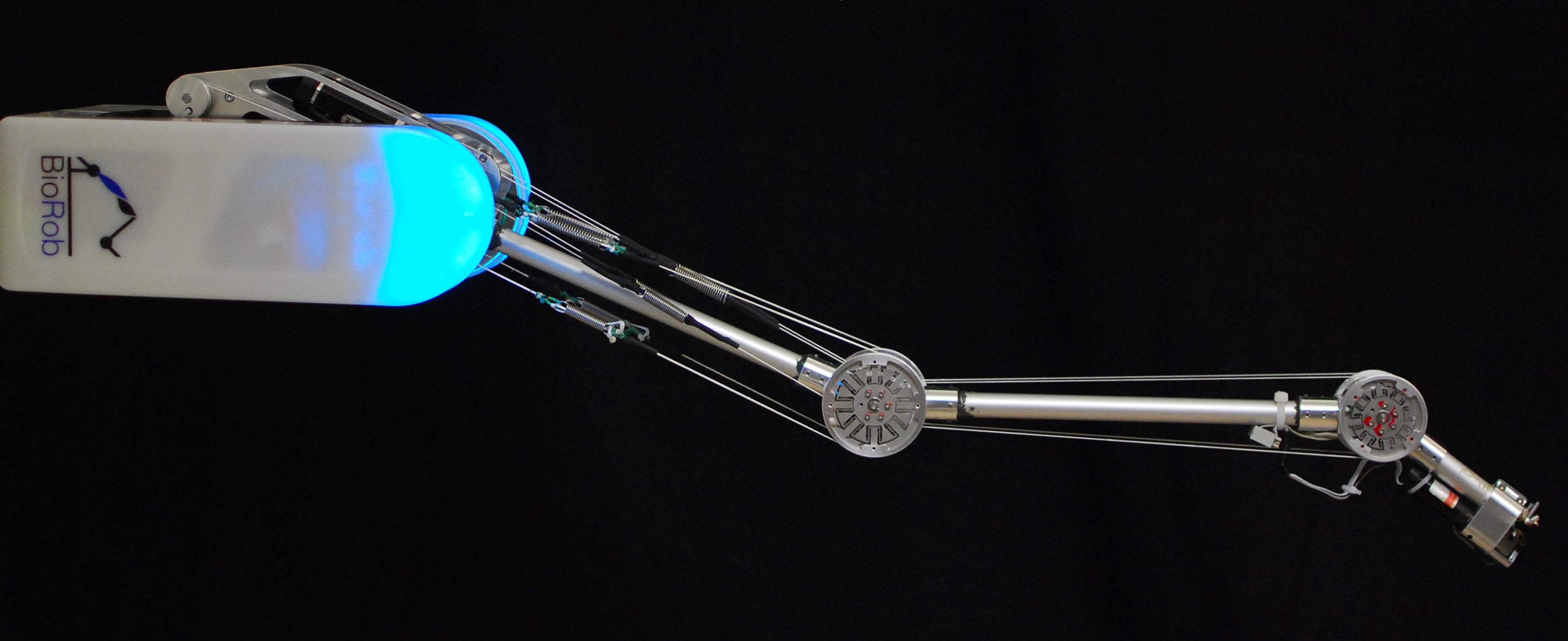}
\caption{Tendon-driven BioRob X4.}
\label{fig:biorob}
\figspace
\end{figure}

Complex robots often violate common modeling assumptions, such as
rigid-body dynamics. A typical example is a tendon-driven robot arm,
shown in Fig.~\ref{fig:biorob}, for which these typical assumption are
violated due to elasticities and springs. Therefore, learning
controllers is a viable alternative to programming robots. To learn
controllers for complex robots, reinforcement learning (RL) is
promising due to the generality of the RL paradigm~\cite{Sutton1998}.
However, without a good initialization (e.g., by human
demonstrations~\cite{Schaal1997,Abbeel2005}) or specific expert
knowledge~\cite{Abbeel2006} RL often relies on data-intensive learning
methods (e.g., Q-learning). For a fragile robotic system, however,
thousands of physical interactions are practically infeasible because
of time-consuming experiments as well as the wear and tear of the robot.

To make RL practically feasible in robotics, we need to speed up
learning by reducing the number of necessary interactions, i.e., robot
experiments. For this purpose, model-based RL is often more promising
than model-free RL, such as Q-learning or
TD-learning~\cite{Atkeson1997a}.  In model-based RL, data is used to
learn a model of the system. This model is then used for policy
evaluation and improvement, reducing the interaction time with the
system. However, model-based RL suffers from \emph{model errors} as it
typically assumes that the learned model closely resembles the true
underlying dynamics~\cite{Schneider1997,Schaal1997}. These model
errors propagate through to the learned policy, whose quality
inherently depends on the quality of the learned model.  A principled
way of accounting for model errors and the resulting optimization bias
is to take the uncertainty about the learned model into account for
long-term predictions and policy learning~\cite{Schneider1997,
  Bagnell2001, Abbeel2006, Ko2007, Deisenroth2011c}.
Besides sample-efficient learning for a single task, generalizing
learned concepts to new situations is a key research topic in RL.
Learned controllers often deal with a single situation\slash context,
e.g., they drive the system to a desired state. In a robotics context,
solutions for multiple related tasks are often desired, e.g., for
grasping multiple objects~\cite{Kroemer2010} or in robot games, such
as learning hitting movements in table tennis~\cite{Mulling2013}, or
in generalizing kicking movements in robot
soccer~\cite{Barrett2010}. 
%

Unlike most other multi-task scenarios, we consider a set-up with a
continuous set of tasks $\vec\task$.  The objective is to learn a
policy that is capable of solving related tasks in the prescribed
class. Since it is often impossible to learn individual policies for
all conceivable tasks, a multi-task learning approach is required that
can generalize across these tasks. We assume that during
\emph{training}, i.e., policy learning, the robot is given a small set
of training tasks $\vec\target_i\train$. In the \emph{test phase}, the
learned policy is expected to generalize from the training tasks to
previously unseen, but related, test tasks $\vec\target\test_i$.

Two general approaches exist to tackle this challenge by either
hierarchically combining local controllers or a richer policy
parametrization.  First, local policies can be learned, and,
subsequently, generalization can be achieved by combining them, e.g.,
by means of a gating network~\cite{Jacobs1991}. This approach has been
successfully applied in RL~\cite{Taylor2007} and
robotics~\cite{Mulling2013}.
In \cite{Mulling2013} a gating network is used to generalize a set of
motor primitives for hitting movements in robot-table tennis. The
limitation of this approach is that it can only deal with convex
combinations of local policies, implicitly requiring local policies
that are linear in the policy parameters.\footnote{One way of making
  hierarchical models more flexible is to learn the hierarchy jointly
  with the local controllers. To the best of our knowledge, such a
  solution does not exist yet.}  In~\cite{Taylor2009,Barrett2010}, it
was proposed to share state-action values across tasks to transfer
knowledge. This approach was successfully applied to kicking a ball
with a NAO robot in the context of RoboCup. However, a mapping from
source to target tasks is explicitly required.  In~\cite{daSilva2012},
it is proposed to sample a number of tasks from a task distribution,
learn the corresponding individual policies, and generalize them to
new problems by combining classifiers and nonlinear
regression. In~\cite{Kober2012,daSilva2012} it is proposed to learn
mappings from tasks to meta-parameters of a policy to generalize
across tasks. The task-specific policies are trained independently,
and the elementary movements are given by Dynamic Movement
Primitives~\cite{Ijspeert2002a}.
Second, instead of learning local policies one can parametrize the
policy directly by the task. For instance, in~\cite{Konidaris2012}, a
value function-based transfer learning approach is proposed that
generalizes across tasks by finding a regression function mapping a
task-augmented state space to expected returns.
We follow this second approach since it allows for generalizing
nonlinear policies:
During training, access to a set of tasks is given and a \emph{single}
controller is learned jointly for all tasks using policy
search. Generalization to unseen tasks in the same domain is achieved
by defining the policy as a function of both the state and the task.
At test time, this allows for generalization to unseen tasks
\emph{without} retraining, which often cannot be done in real time.
%
For learning the parameters of the multi-task policy, we use the
\textsc{pilco} policy search framework~\cite{Deisenroth2014}.
\textsc{Pilco} learns flexible Gaussian process (GP) forward models
and uses fast deterministic approximate inference for long-term
predictions to achieve data-efficient learning. In a robotics context,
policy search methods have been successfully applied to many
tasks~\cite{Deisenroth2013} and seem to be more promising than value
function-based methods for learning policies. Hence, this paper
addresses two key problems in robotics: multi-task and
data-efficient policy learning.



\section{Policy Search for Learning Multiple Tasks}
\label{sec:spmt}

We consider dynamical systems
%
$
\vec x_{t+1} = f(\vec x_{t},\vec u_{t})+\vec w
$
%
with continuous states $\vec x\in\R^D$ and controls $\vec u\in\R^F$
and unknown transition dynamics $f$. The term $\vec w\sim\gauss{\vec
  0}{\mat\Sigma_w}$ is zero-mean i.i.d. Gaussian noise with covariance
matrix $\mat\Sigma_w$. In (single-task) policy search, our objective
is to find a deterministic \emph{policy} $\pi:\vec x\mapsto \pi(\vec
x,\vec\polpar)=\vec u$ that minimizes the \emph{expected long-term
  cost}
\begin{align}
  \label{eq:expected return}
  \hspace{-2mm}J^\pi(\vec\polpar) = \sum\nolimits_{t = 1}^T\E_{\vec x_t}[\cost(\vec
  x_t)]\,,\quad p(\vec x_0) = \gauss{\vec\mu_0^x}{\mat\Sigma_0^x}\,,
\end{align}
of following $\pi$ for $T$ steps. Note that the trajectory $\vec
x_1,\dotsc, \vec x_T$ depends on the policy $\pi$ and, thus, the
parameters $\vec\polpar$.  In Eq.~(\ref{eq:expected return}),
$\cost(\vec x_t)$ is a given cost function of state $\vec x$ at time
$t$. The policy $\pi$ is parametrized by
$\vec\polpar\in\R^P$. Typically, the cost function $c$ incorporates
some information about a task $\vec\target$, e.g., a desired target
location $\vec x_{\text{target}}$ or a trajectory. Finding a policy
that minimizes Eq.~(\ref{eq:expected return}) solves the task
$\vec\task$ of controlling the robot toward the target.

\subsection{Task-Dependent Policies}

We propose to learn a \emph{single} policy for all tasks jointly to
generalize classical policy search to a multi-task scenario. We assume
that the dynamics are stationary with the transition probabilities and
control spaces shared by all tasks. By learning a single policy that
is sufficiently flexible to learn the training tasks
$\vec\task_i\train$, we aim to obtain good generalization performance
to related test tasks $\vec\task_j\test$ by reducing the danger of
overfitting to the training tasks, a common problem with current
hierarchical approaches.

To learn a single controller for multiple tasks $\vec\task_k$, we
propose to make the policy a function of the state $\vec x$, the
parameters $\vec\polpar$, \emph{and} the task $\vec\task$, such that
$\vec u = \pi(\vec x,\vec\task, \vec\polpar)$. In this way, a trained
policy has the potential to generalize to previously unseen tasks by
computing different control signals for a fixed state $\vec x$ and
parameters $\vec\theta$ but varying tasks $\vec\task_k$.
\begin{figure}
\centering
\includegraphics[width =0.8\hsize]{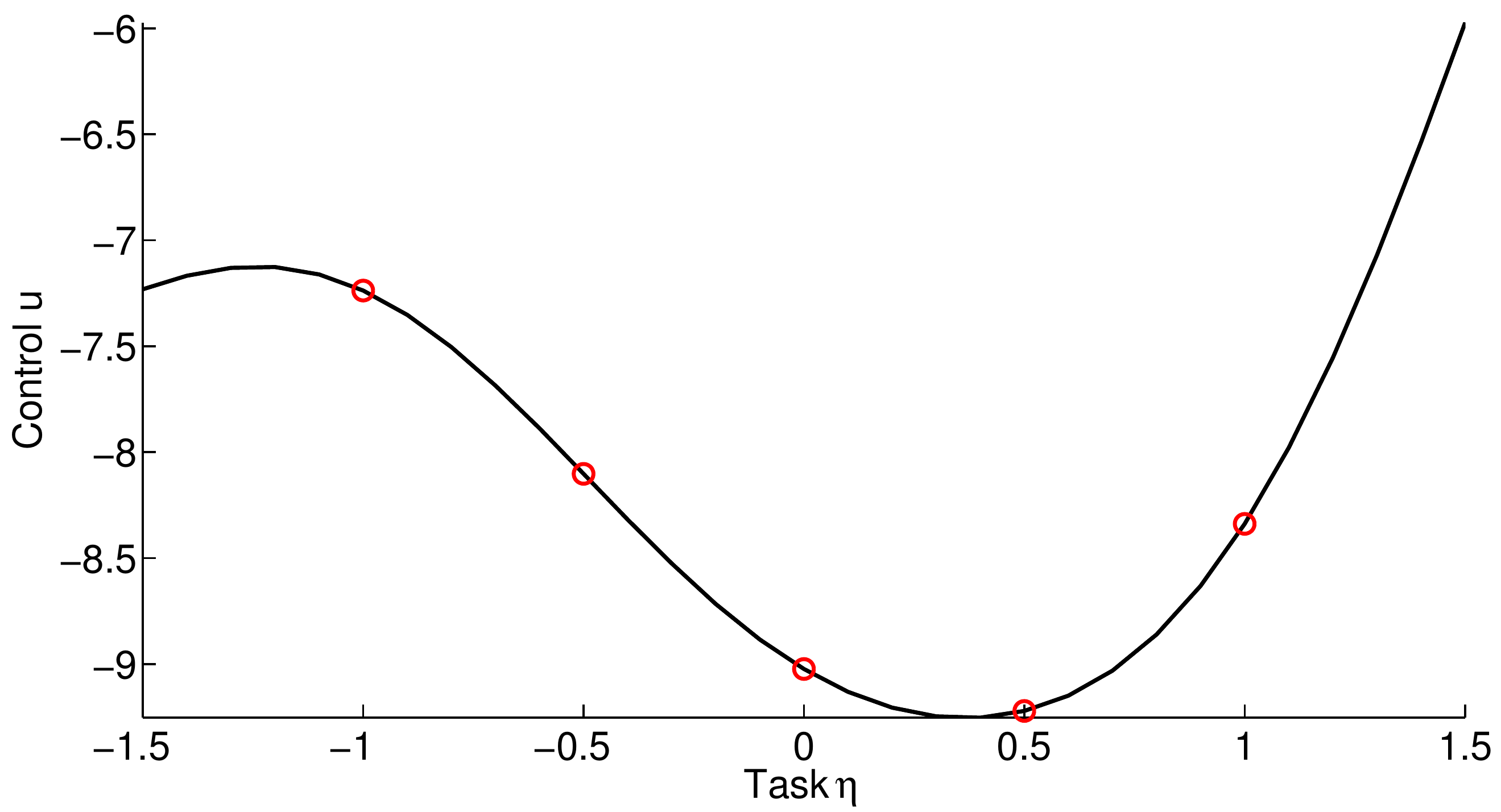}
\caption{Generalization ability of a multi-task policy for the
  cart-pole experiment in Sec.~\ref{sec:cp}. Here, the
  state is fixed, the change in the controls is solely due to the
  change in the task.  The black line represents the corresponding
  policy that has been augmented with the task. The controls of the
  training tasks are denoted by the red circles. The policy smoothly
  generalizes across test tasks.}
\label{fig:generalization_motivation}
\end{figure}
Fig.~\ref{fig:generalization_motivation} gives an intuition of what
kind of generalization power we can expect from a policy that uses
state-task pairs as inputs: Assume a given policy parametrization, a
fixed state, and five training targets $\vec\target_i\train$. For each
pair $(\vec x,\vec\target_i\train, \vec\polpar)$, the policy
determines the corresponding controls $\pi(\vec
x,\vec\target_i\train,\vec\polpar)$, which are denoted by the red
circles. The differences in these control signals are achieved solely
by  changing $\vec\target_i\train$ in $\pi(\vec
x,\vec\target_i\train,\vec\polpar)$ as $\vec x$ and $\vec\polpar$ were
assumed fixed. The parametrization of the policy by $\vec\polpar$ and
$\vec\task$ implicitly determines the generalization power of $\pi$ to
new (but related) tasks $\vec\task_j\test$ at test time. The policy
for a fixed state but varying test tasks $\vec\task_j\test$ is
represented by the black curve.
To find good parameters of the multi-task policy, we incorporate our
multi-task learning approach into the model-based \textsc{pilco}
policy search framework~\cite{Deisenroth2011c}.  The high-level steps
of the resulting algorithm are summarized in
Fig.~\ref{alg:pilco}.
%
\begin{figure}[h]
\fcolorbox{white}{lightgray}{\parbox{\hsize}{%
   \color{black}%
\caption{Multi-Task Policy Search}
\label{alg:pilco}
\begin{algorithmic}[1]
  \STATE {\bfseries init:} Pass in training tasks $\vec\task_i\train$,
  initialize policy parameters $\vec\polpar$ randomly. Apply random
  control signals and record data.
   \label{alg:random controls}
   \REPEAT
   \STATE Update GP forward dynamics model using all data
\label{alg:model learning}
\REPEAT \label{alg:line policy search} \STATE Long-term predictions:
Compute $\E_{\vec\target}[J^\pi(\vec\polpar,\vec\task)]$
  \label{alg:line policy evaluation}
  \STATE Analytically compute gradient $\E_{\vec\task}[d
  J^\pi(\vec\polpar,\vec\task)\slash d\vec\polpar]$
\label{alg:line get gradient}
\STATE Update policy parameters $\vec\polpar$ (e.g., BFGS)
\label{alg:update parameters}
  \UNTIL{convergence; {\bf return} $\vec\polpar^*$}
  \STATE Set $\pi^*\leftarrow \pi(\vec\polpar^*)$
  \STATE Apply $\pi^*$ to robot and record data
  \label{alg:line apply controller}
  \UNTIL{$\forall i$: task $\vec\task_i\train}$ learned 
  \STATE Apply $\pi^*$ to test tasks $\vec\task\test_j$
  \label{alg:line test policy}
\end{algorithmic}
}}
\end{figure}
We assume that a set of training tasks $\vec\task\train_i$ is given.
The parametrized policy $\pi$ is initialized randomly, and,
subsequently, applied to the robot, see line~\ref{alg:random
  controls} in Fig.~\ref{alg:pilco}. Based on the initial collected
data, a probabilistic GP forward model of the underlying robot
dynamics is learned (line~\ref{alg:model learning}) to consistently
account for model errors~\cite{Rasmussen2006}.

We define the policy $\pi$ as an explicit function of both the
state $\vec x$ and the task $\vec\task$, which essentially means that
the policy depends on a task-augmented state and $\vec u = \pi(\vec x,
\vec\task,\vec\polpar)$. 
Before going into detail, let us consider the case where a function $g$
relates state and task. In this paper, we consider two cases: (a) A
linear relationship between the task $\vec\task$ and the state $\vec
x_t$ with $g(\vec x_t, \vec\task) = \vec\task - \vec x_t$. For
example, the state and the task (corresponding to a target location)
can be both defined in camera coordinates, and the target location
parametrizes and defines the task. (b) The task variable $\vec \task$
and the state vector are not directly related, in which case $g(\vec
x_t, \vec \task)= \vec \task$. For instance, the task variable could
simply be an index.
We approximate the joint distribution $p(\vec x_t, g(\vec
x_t,\vec\task))$ by a Gaussian
\begin{align}
  \mathcal N\left(
\begin{bmatrix}
\vec\mu_t^x\\
\vec\mu_t^\target
\end{bmatrix}
,\,
\begin{bmatrix}
\mat\Sigma_t^x & \mat C_t^{x\target}\\
\mat C_t^{\target x} & \mat\Sigma_t^\target
\end{bmatrix}
\right)\eqqcolon\gaussx{\vec
  x_t^{x,\target}}{\vec\mu_t^{x,\target}}{\mat\Sigma_t^{x,\target}}\,,
\label{eq:augmented state distribution}
\end{align}
where the state distribution is
$\gaussx{\vec x_t}{\vec\mu_t^x}{\mat\Sigma_t^x}$ and $\mat
C_t^{x\target}$ is the cross-covariance between the state and $g(\vec
x_t, \vec\target)$.
%
The cross-covariances for $g(\vec x_t, \vec\target) = \vec\target
-\vec x_t$ are $\mat C_t^{x\target} = -\mat\Sigma_t^x$. If the state
and the task are not directly related, i.e., $g(\vec x_t, \vec\task) =
\vec\task$, then $\mat C_t^{x\target}=\mat 0$.

The Gaussian approximation of the joint distribution $p(\vec x_t,
g(\vec x_t,\vec\target))$ in Eq.~\eqref{eq:augmented state
  distribution} serves as the input distribution to the controller
function $\pi$. Although we assume that the tasks $\vec\target_i\test$
are given deterministically at test time, introducing a task
uncertainty $\mat\Sigma_t^\target > \mat 0$ during \emph{training} can
make sense for two reasons: First, during \emph{training}
$\mat\Sigma_t^\target$ defines a \emph{task distribution}, which may
allow for better generalization performance compared to
$\mat\Sigma_t^\target = \mat 0$. Second, $\mat\Sigma_t^\target > \mat
0$ induces uncertainty into planning and policy learning. Therefore,
$\mat\Sigma_t^\target$ serves as a regularizer and makes policy
overfitting less likely.

\subsection{Multi-Task Policy Evaluation}
For policy evaluation, we analytically approximate the expected
long-term cost $J^\pi(\vec\polpar)$ by averaging over all tasks
$\vec\task$, see line~\ref{alg:line policy evaluation} in
Fig.~\ref{alg:pilco}, according to
\begin{align}
  \E_{\vec\target} [J^\pi(\vec\polpar,\vec\task)] & \approx\tfrac{1}{M}
  \sum\nolimits_{i = 1}^{M} J^\pi(\vec\polpar,\vec\target_i\train)\,,
\label{eq:new J-function}
\end{align}
where $M$ is the number of tasks considered during training.  The
expected cost $J^\pi(\vec\polpar,\vec\target_i\train)$ corresponds to
Eq.~\eqref{eq:expected return} for a specific training task
$\vec\target_i\train$.  The intuition behind the expected long-term
cost in Eq.~(\ref{eq:new J-function}) is to allow for learning a
single controller for multiple tasks \emph{jointly}. Hence, the
controller parameters $\vec\polpar$ have to be updated in the context
of \emph{all} tasks. The resulting controller is not necessarily
optimal for a \emph{single} task, but (neglecting approximations and
local minima) optimal across \emph{all tasks} on average, presumably
leading to good generalization performance. The expected long-term
cost $J^\pi(\vec\polpar,\vec\task\train_i)$ in Eq.~(\ref{eq:new
  J-function}) is computed as follows.

First, based on the learned GP dynamics model, approximations to the
long-term predictive state distributions $p(\vec x_1|\vec\task),
\dotsc, p(\vec x_T|\vec\task)$ are computed analytically: Given a
joint Gaussian prior distribution $\prob(\vec x_{t},\vec
u_{t}|\vec\task)$, the distribution of the successor state
\begin{align}
  \prob(\vec x_{t+1}|\vec\task\train_i)\! =\! \iiint\!\prob(\vec
  x_{t+1}|\vec x_{t},\vec u_{t})\prob(\vec x_{t},\vec
  u_{t}|\vec\task_i\train)d f d\vec x_{t}d\vec u_{t}
\label{eq:next-state integrals}
\end{align}
cannot be computed analytically for nonlinear covariance
functions. However, we approximate it by a Gaussian distribution
$\gaussx{\vec x_{t+1}}{\vec\mu_{t+1}^x}{\mat\Sigma_{t+1}^x}$ using
exact moment matching~\cite{Quinonero-Candela2003a,
  Deisenroth2014}. In Eq.~\eqref{eq:next-state integrals}, the
transition probability $ \prob(\vec x_{t+1}|\vec x_{t},\vec
u_{t})=\prob(f(\vec x_{t}, \vec u_{t})|\vec x_{t},\vec u_{t})$ is the
GP predictive distribution at $(\vec x_{t},\vec u_{t})$. Iterating the
moment-matching approximation of Eq.~(\ref{eq:next-state integrals})
for all time steps of the finite horizon $T$ yields Gaussian marginal
predictive distributions $\prob(\vec
x_1|\vec\task_i\train),\dotsc,\prob(\vec x_T|\vec\task_i\train)$.

Second, these approximate Gaussian long-term predictive state
distributions allow for the computation of the expected immediate cost
$ \E_{\vec x_t}[c(\vec x_t)|\vec\task_i\train] = \int c_\task(\vec
x_t)p(\vec x_t|\vec\task_i\train)d\vec x_t $
for a particular task $\vec\task_i\train$, where $p(\vec
x_t|\vec\task_i\train) = \gaussx{\vec
  x_t}{\vec\mu_t^x}{\mat\Sigma_t^x}$ and $c_\task$ is a task-specific
cost function.  This integral can be solved analytically for many
choices of the immediate cost function $c_\task$, such as polynomials,
trigonometric functions, or unnormalized Gaussians. Summing the values
$\E_{\vec x_t}[c(\vec x_t)|\vec\task_i\train]$ from $t=1,\dotsc, T$
finally yields $J^\pi(\vec\polpar,\vec\task_i\train)$ in
Eq.~(\ref{eq:new J-function}).

\subsection{Gradient-based Policy Improvement}
\label{sec:pilco derivatives}
The deterministic and analytic approximation of
$J^\pi(\vec\polpar,\vec\task)$ by means of moment matching allows for
an analytic computation of the corresponding gradient $d
J^\pi(\vec\polpar,\vec\task)/d\vec\polpar$ with respect to the policy
parameters $\vec\polpar$, see Eq.~(\ref{eq:new J-function}) and
line~\ref{alg:line get gradient} in Fig.~\ref{alg:pilco}, which are
given by
\begin{align}
  {\footnotesize \frac{d J^\pi(\vec\polpar,\vec\task)}{d\vec\polpar}} &=
  \sum\nolimits_{t = 1}^T\ffrac{d}{d\vec\polpar}\E_{\vec x_t}[c(\vec
  x_t)|\vec\task]\,.
\label{eq:top-level derivative}
\end{align}
These gradients can be used in any gradient-based optimization
toolbox, e.g., BFGS (line~\ref{alg:update parameters}). Analytic
computation of $J^\pi(\vec\polpar,\vec\task)$ and its gradients $d
J^\pi(\vec\polpar,\vec\task)/d\vec\polpar$ is more efficient than
estimating policy gradients through sampling: For the latter, the
variance in the gradient estimate grows quickly with the number of
policy parameters and the horizon $T$~\cite{Peters2006}.

Computing the derivatives of $J^\pi(\vec\polpar,\vec\task_i\train)$
with respect to the policy parameters $\vec\polpar$ requires repeated
application of the chain-rule.  Defining $\mathcal E_t\coloneqq
\E_{\vec x_t}[c(\vec x_t)|\vec\task_i\train]$ in
\eq~\eqref{eq:top-level derivative} yields
\begin{align}
  \ffrac{d\mathcal E_t}{d\vec\polpar}&= \ffrac{d\mathcal
    E_t}{d\prob(\vec x_t)}\ffrac{d\prob(\vec x_t)}{d\vec\polpar}
  \coloneqq \ffrac{\partial\mathcal E_t
  }{\partial\vec\mu_t^x}\ffrac{d\vec\mu_t^x}{d\vec\polpar}+
  \ffrac{\partial\mathcal E_t
  }{\partial\mat\Sigma_t^x}\ffrac{d\mat\Sigma_t^x}{d\vec\polpar}\,,
  \label{eq:chain rule}
\end{align} 
where we took the derivative with respect to $\prob(\vec x_t)$, i.e.,
the parameters of the state distribution $\prob(\vec x_t)$. In
Eq.~(\ref{eq:chain rule}), this amounts to computing the derivatives
of $\mathcal E_t$ with respect to the mean $\vec\mu_t^x$ and
covariance $\mat\Sigma_t^x$ of the Gaussian approximation of
$\prob(\vec x_t)$.
The chain-rule yields the total derivative of $p(\vec x_t)$ with
respect to $\vec\polpar$
\begin{align}
\ffrac{d\prob(\vec x_t)}{d\vec\polpar} & =
 \ffrac{\partial p(\vec x_t)}{\partial p(\vec x_{t-1})}
\ffrac{d\prob(\vec
    x_{t-1})}{d\vec\polpar} + \ffrac{\partial\prob(\vec
    x_t)}{\partial\vec\polpar}\,.\label{eq:dp(x)dpsi}
\end{align}
%
%
In Eq.~\eqref{eq:dp(x)dpsi}, we assume that the total derivative
$d\prob(\vec x_{t-1})/d\vec\polpar$ is known from the computation for
the previous time step. Hence, we only need to compute the partial
derivative $\partial\prob(\vec x_t)/\partial\vec\polpar$. Note that
$\vec x_t = f(\vec x_{t-1}, \vec u_{t-1}) + \vec w$ and $\vec u_{t-1}
= \pi(\vec x_{t-1}, g(\vec x_{t-1},\vec\eta), \vec\theta)$. Therefore,
we obtain, with the Gaussian approximation to the marginal state
distribution $p(\vec x_t)$, $\partial\prob(\vec
x_t)/\partial\vec\polpar = \{\partial\vec\mu_t^x/\partial\vec\polpar,
\partial\mat\Sigma_t^x/\partial\vec\polpar\}$
with
\begin{align}
  \hspace{-2mm}\ffrac{\partial\{\vec\mu_t^x,\mat\Sigma_t^x\}}{\partial\vec\polpar}
  &\!=\!
  \ffrac{\partial\{\vec\mu_t^x,\mat\Sigma_t^x\}}{\partial\prob(\vec
    u_{t-1})}\ffrac{\partial\prob(\vec
    u_{t-1})}{\partial\vec\polpar}\nonumber\\
&\!=\!\ffrac{ \partial\{\vec\mu_t^x,\mat\Sigma_t^x\}}{\partial\vec\mu_{t-1}^u}
  \red{\ffrac{\partial\vec\mu_{t-1}^u}{\partial\vec\polpar}}\!+\!
  \ffrac{\partial\{\vec\mu_t^x,\mat\Sigma_t^x\}}{\partial\mat\Sigma_{t-1}^u}
  \red{\ffrac{\partial\mat\Sigma_{t-1}^u}{\partial\vec\polpar}}.
 \label{eq:dmu-Sigma/dtheta}
\end{align}
Here, the distribution $$\prob(\vec u_{t-1})=\int\pi(\vec
x_{t-1},g(\vec x_{t-1},\vec\task),\vec\polpar)p(\vec x_{t-1}) d\vec
x_{t-1}$$ of the control signal is approximated by a Gaussian with mean
$\vec\mu_{t-1}^u$ and covariance $\mat\Sigma_{t-1}^u$. These moments
(and their gradients with respect to $\vec\polpar$) can often be
computed analytically, e.g., in linear models with polynomial or
Gaussian basis functions.
%
The augmentation of the policy with the (transformed) task variable
requires an additional layer of gradients for computing $d
J^\pi(\vec\polpar)/d\vec\polpar$. The variable transformation affects
the partial derivatives of $\vec\mu_{t-1}^u$ and $\mat\Sigma_{t-1}^u$
(marked red in Eq.~(\ref{eq:dmu-Sigma/dtheta})), such that
\begin{align}
  \red{\ffrac{\partial\{\vec\mu_{t-1}^u,\mat\Sigma_{t-1}^u\}}{\partial\{\vec\mu_{t-1}^x,\mat\Sigma_{t-1}^x,\vec\theta\}}}
  &= \ffrac{\partial\{\vec\mu_{t-1}^u,\mat\Sigma_{t-1}^u\}}{\partial\prob(\vec
    x_{t-1}, g(\vec x_{t-1},\vec\target)) }\nonumber\\
&\quad\times{\ffrac{\partial\prob(\vec x_{t-1},
      g(\vec x_{t-1},\vec\target)) }{\partial\{\vec\mu_{t-1}^x,\mat\Sigma_{t-1}^x,\vec\theta\}}}\,,
\label{eq:first additional derivative}
\end{align}
which can often be computed analytically. Similar
to~\cite{Deisenroth2014}, we combine these derivatives with the
gradients in Eq.~(\ref{eq:dmu-Sigma/dtheta}) via the chain and
product-rules, yielding an analytic gradient $d
J^\pi(\vec\polpar,\vec\task_i\train)/d\vec\polpar$ in
Eq.~\eqref{eq:new J-function}, which is used for gradient-based policy
updates, see lines~\ref{alg:line get gradient}--\ref{alg:update
  parameters} in Fig.~\ref{alg:pilco}.



\section{Evaluations and Results}

In the following, we analyze our approach to multi-task policy search
on three scenarios: 1) the under-actuated cart-pole swing-up
benchmark, 2) a low-cost robotic manipulator system that learns block
stacking, and 3) an imitation learning ball-hitting task with a
tendon-driven robot. In all cases, the system dynamics were unknown
and inferred from data using GPs.

\subsection{Multi-Task Cart-Pole Swing-up}
\label{sec:cp}
We applied our proposed multi-task policy search to learning a model
and a controller for the cart-pole swing-up.
The system consists of a cart with mass $\unit[0.5]{kg}$ and a
pendulum of length $\unit[0.6]{m}$ and mass $\unit[0.5]{kg}$
attached to the cart. Every $\unit[0.1]{s}$, an external force was
applied to the cart, but not to the pendulum. The friction between
cart and ground was $\unit[0.1]{Ns/m}$.
%
The state $\vec x=[\chi, \dot \chi, \varphi, \dot\varphi]$ of the
system comprised the position $\chi$ of the cart, the velocity $\dot
\chi$ of the cart, the angle $\varphi$ of the pendulum, and the
angular velocity $\dot\varphi$ of the pendulum. For the equations of
motion, we refer to~\cite{Deisenroth2010b}.  The nonlinear controller
was parametrized as a regularized RBF network with 100 Gaussian basis
functions. The controller parameters were the locations of the basis
functions, a shared (diagonal) width-matrix, and the weights,
resulting in approximately 800 policy parameters.

Initially, the system was expected to be in a state, where the
pendulum hangs down; more specifically, $p(\vec x_0) = \gauss{\vec 0}{
  0.1^2\mat I}$. By pushing the cart to the left and to the right, the
objective was to swing the pendulum up and to balance it in the
inverted position at a target location $\target$ of the cart specified
at \emph{test time}, such that $\vec x_{\text{target}} =
[\target,*,\pi+2k\pi,*]$ with $\target\in[-1.5,1.5]\,\unit{m}$ and
$k\in\Z$. The cost function $c$ in Eq.~(\ref{eq:expected return}) was
chosen as $c(\vec x) = 1-\exp(-8\|\vec x - \vec
x_{\text{target}}\|^2)\in[0,1]$ and penalized the Euclidean distance
of the tip of the pendulum from its desired inverted position with the
cart being at target location $\target$.  Optimally solving the task
required the cart to stop at the target location
$\vec\target$. Balancing the pendulum with the cart offset by
$\unit[20]{cm}$ caused an immediate cost (per time step) of about 0.4.
We considered four experimental setups:

\noindent\textbf{Nearest neighbor independent controllers (NN-IC):}
Nearest-neighbor baseline experiment with five independently learned
controllers for the desired swing-up locations $\eta=\{\unit[\pm
1]{m}, \unit[\pm 0.5]{m}, \unit[0]{m}\}$. Each controller was learned
using the \textsc{pilco} framework~\cite{Deisenroth2011c} in 10 trials
with a total experience of $\unit[200]{s}$. For the test tasks
$\vec\task\test$, we applied the controller with the closest training
task $\vec\target\train$.

\noindent\textbf{Re-weighted independent controllers (RW-IC):}
Training was identical to NN-IC. At test time, we combined individual
controllers using a gating network, similar to~\cite{Mulling2013},
resulting in a convex combination of local policies. The
gating-network weights were
\begin{align}
 v_i = \frac{\exp\big(-\tfrac{1}{2\kappa}\|\vec\task\test -
    \vec\task_i\train\|^2\big)}{{\sum\nolimits_j^{|\vec\task\train|}
    \exp\big(-\tfrac{1}{2\kappa}\|\vec\task\test -
    \vec\task_j\train\|^2\big)}}\,,
\label{eq:gating network weights}
\end{align}
such that the applied control signal was $\vec u = \sum\nolimits_i
v_i\pi_i(\vec x)$. An extensive grid search resulted in
$\kappa=\unit[0.0068]{m^2}$, leading to the best test performance in
this scenario, making RW-IC nearly identical to the NN-IC.

\noindent\textbf{Multi-task policy search, $\mat\Sigma^\task=\mat 0$
  (MTPS0):} 
Multi-task policy search with five known tasks during training, which
only differ in the location of the cart where the pendulum is supposed
to be balanced. The target locations were $\task\train=\{\unit[\pm
1]{m}, \unit[\pm 0.5]{m}, \unit[0]{m}\}$. Moreover, $g(\vec x_t,
\task) = \task -\chi(t)$ and $\mat\Sigma^\task=\mat 0$. We show
results after 20 trials, i.e., a total experience of $\unit[70]{s}$
only.

\noindent\textbf{Multi-task policy search, $\mat\Sigma^\task>\mat 0$
  (MTPS+):} 
Multi-task policy search with the five training tasks
$\task\train=\{\unit[\pm 1]{m}, \unit[\pm 0.5]{m}, \unit[0]{m}\}$, but
with training task covariance $\mat\Sigma^\task = \diag([0.1^2, 0, 0,
0])$. We show results after 20 trials, i.e., $\unit[70]{s}$ total
experience.

For testing the performance of the algorithms, we applied the learned
policies 100 times to the test-target locations $\vec\task\test =
-1.5, -1.4, \dotsc, 1.5$. Every time, the initial state of a rollout
was sampled from $\prob(\vec x_0)$. For the MTPS experiments, we
plugged the test tasks into Eq.~\eqref{eq:augmented state
  distribution} to compute the corresponding control signals.

\begin{figure*}[tb]
\subfigure[NN-IC: No generalization.]{
  \includegraphics[width =
  \twofig]{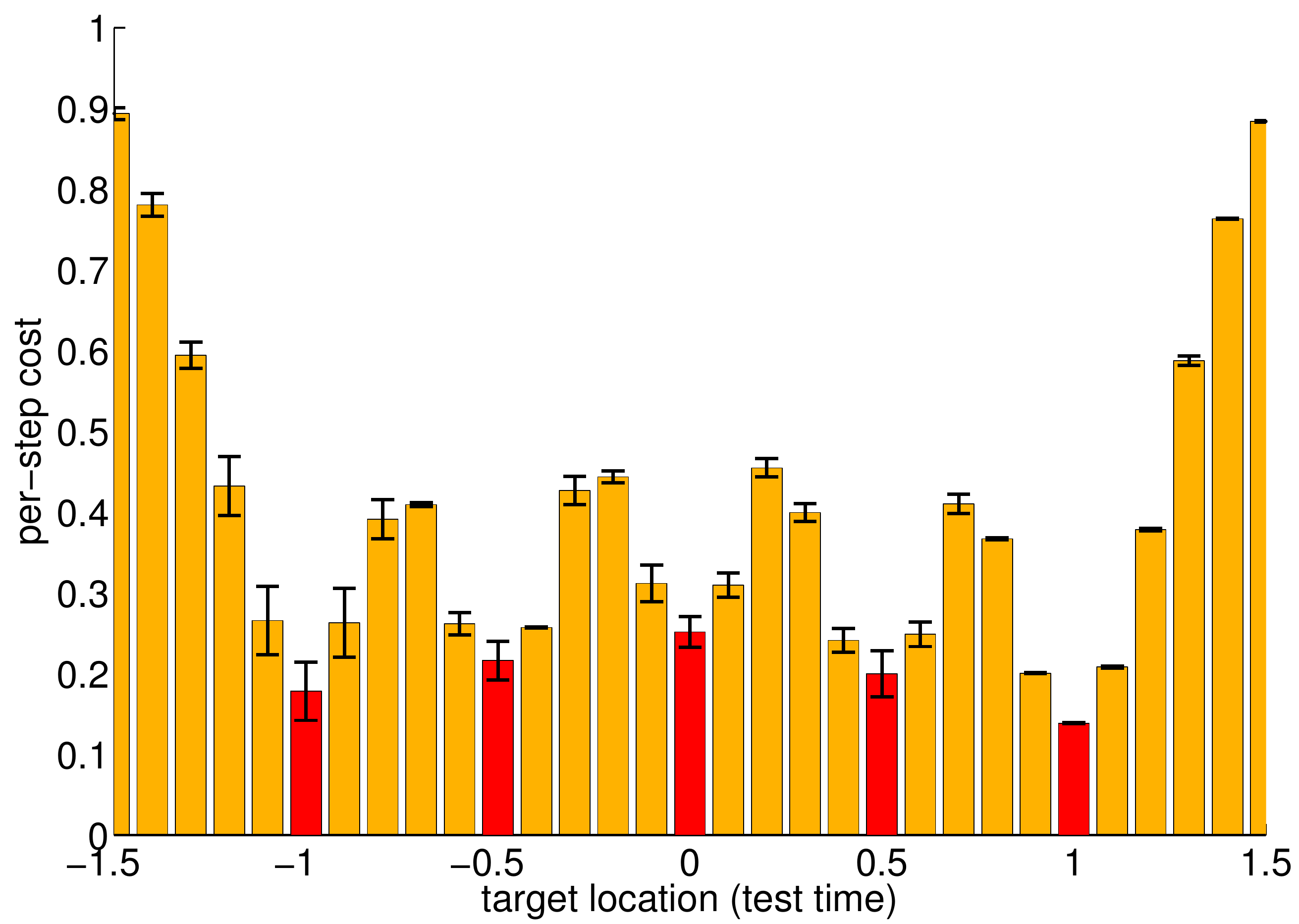}
\label{fig:cp_generalization individual controllers}
}
\hfill
\subfigure[RW-IC: No generalization.]{
  \includegraphics[width =
  \twofig]{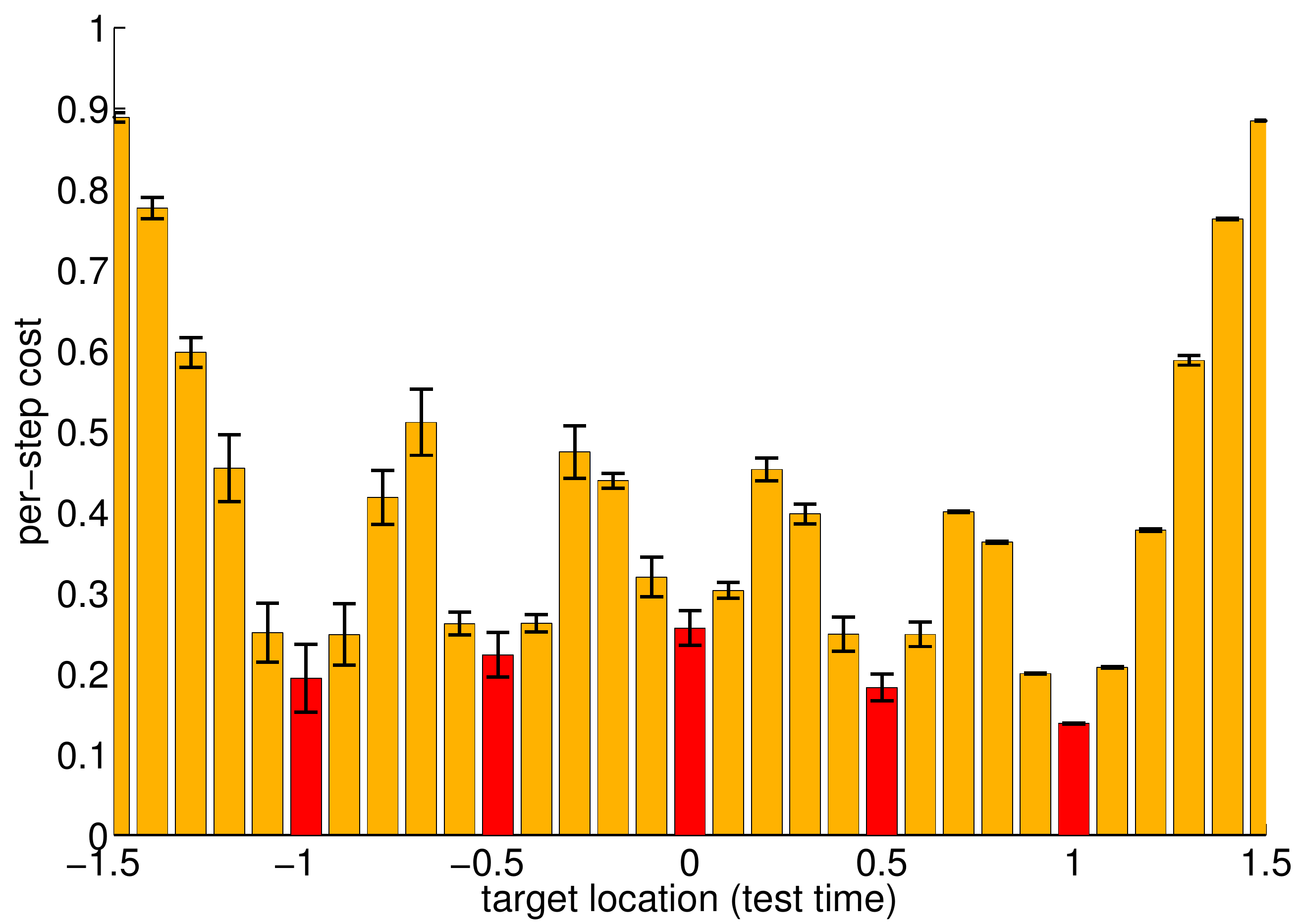}
\label{fig:cp_generalization hierarchical controller}
}
\hfill
\subfigure[MTPS0: Some generalization.]{
\includegraphics[width =
\twofig]{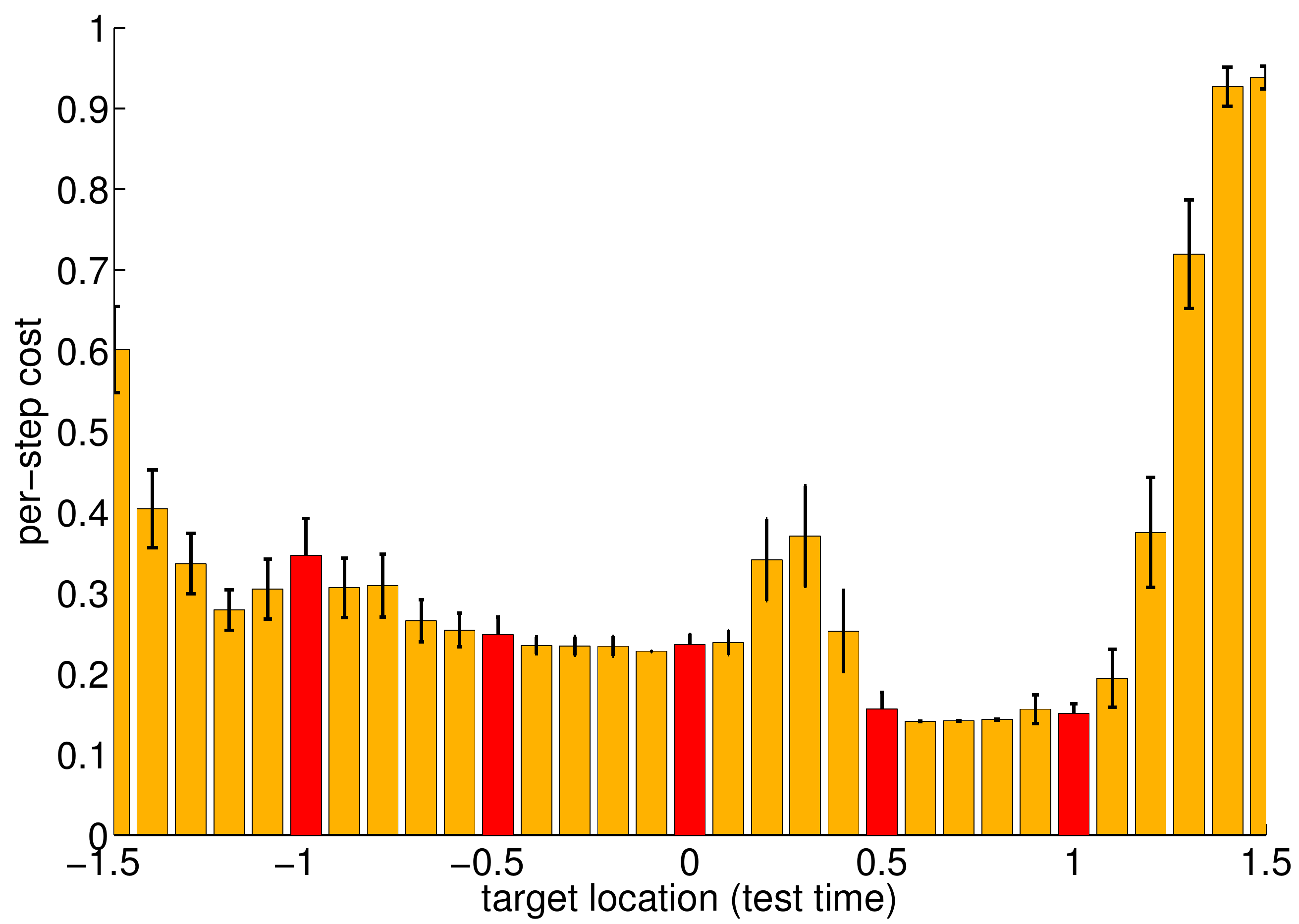} 
\label{fig:cp_generalization spmt deterministic}
} \hfill \subfigure[MTPS+: Good generalization in the ``coverage'' of
the Gaussians.]{
  \includegraphics[width =
  \twofig]{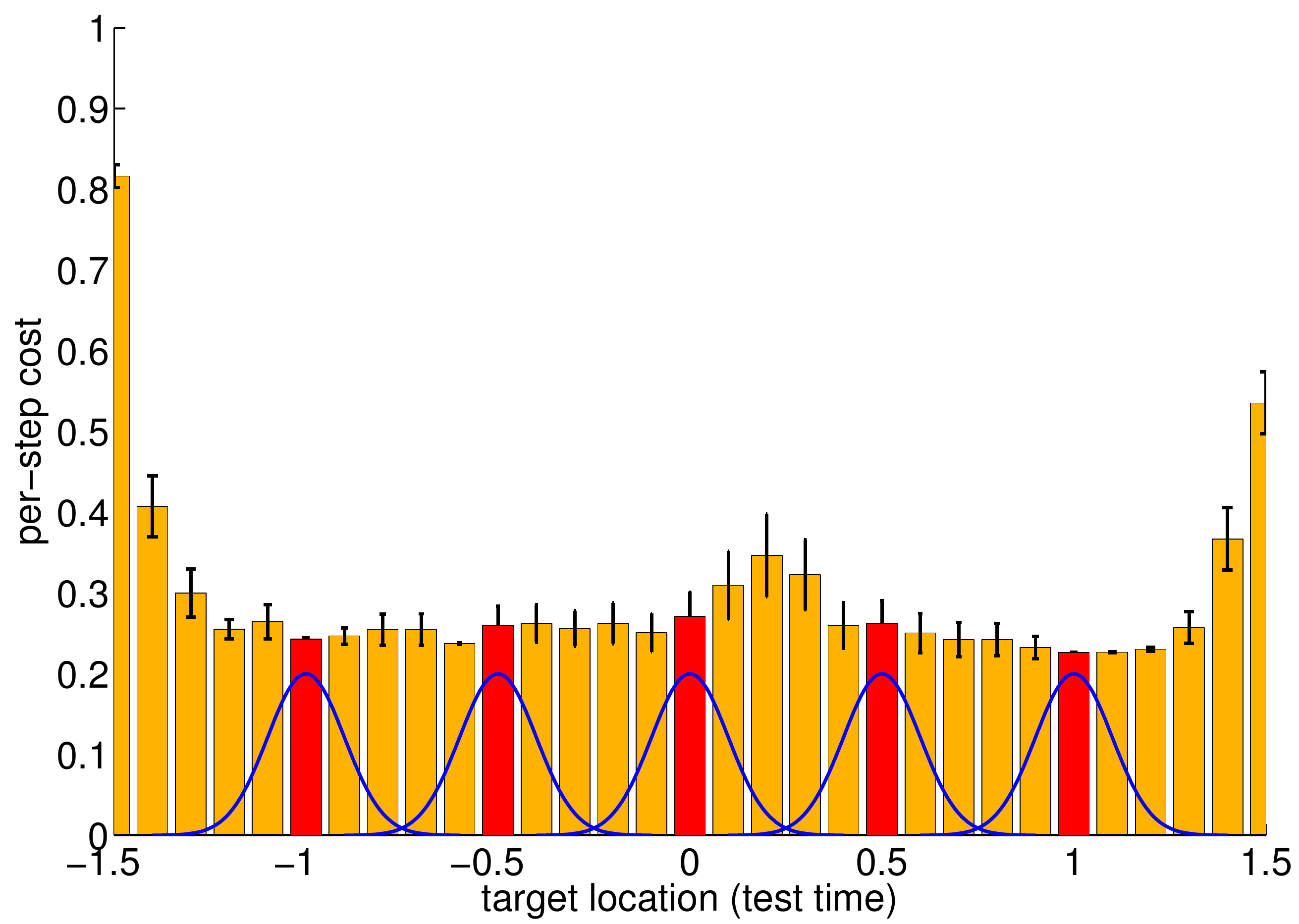}
\label{fig:cp_generalization spmt uncertain}
}
\caption{Generalization performance for the multi-task
  cart-pole swing-up. The graphs show the expected cost
  per time step along with twice the standard errors.}
\label{fig:cp_generalizations}
\figspace
\end{figure*}
Fig.~\ref{fig:cp_generalizations} illustrates the generalization
performance of the learned controllers. The horizontal axes denote the
locations $\task\test$ of the target position of the cart at
\emph{test} time. The height of the bars show the average (over
trials) cost per time step. The means of the \emph{training} tasks
$\target\train$ are the location of the red bars. For Experiment 1,
\fig~\ref{fig:cp_generalization spmt uncertain} shows the distribution
$p(\task_i\train)$ used during training as the bell-curves, which
approximately covers the range
$\task\in[\unit[-1.2]{m},\unit[1.2]{m}]$.

The NN-IC controller in the nearest-neighbor baseline, see
Fig.~\ref{fig:cp_generalization individual controllers}, balanced the
pendulum at a cart location that was not further away than
$\unit[0.2]{m}$, which incurred a cost of up to 0.45.
In Fig.~\ref{fig:cp_generalization hierarchical controller}, the
performances for the hierarchical RW-IC controller are shown. The
performance for the best value $\kappa$ in the gating network, see
Eq.~(\ref{eq:gating network weights}), was similar to the performance
of the NN-IC controller. However, between the training tasks for the
local controllers, where the test tasks were in the range of
$[\unit[-0.9]{m},\unit[0.9]{m}]$, the convex combination of local
controllers led to more failures than in NN-IC, where the pendulum
could not be swung up successfully: Convex combinations of nonlinear
local controllers eventually decreased the (non-existing)
generalization performance of RW-IC.

Fig.~\ref{fig:cp_generalization spmt deterministic} shows the
performance of the MTPS0 controller. The
MTPS0 controller successfully performed the swing-up
plus balancing task for all tasks $\task\test$ close to the training
tasks.  However, the performance varied relatively strongly.
Fig.~\ref{fig:cp_generalization spmt uncertain} shows that the MTPS+
controller successfully performed the swing-up plus balancing task for
all tasks $\task\test$ at test time that were sufficiently covered by
the uncertain training tasks $\task\train_i, i = 1,\dotsc,5$,
indicated by the bell curves representing
$\mat\Sigma^\task>0$. Relatively constant performance across the test
tasks covered by the bell curves was achieved. An average cost of 0.3
meant that the pendulum might be balanced with the cart slightly
offset. Fig.~\ref{fig:generalization_motivation} shows the learned
MTPS+ policy for all test tasks $\task\test$ with the state $\vec x =
\vec \mu_0$ fixed.

\begin{table}[b]
\vspace{-5mm}
\caption{Multi-task cart-pole swing-up: Average costs across  31 test tasks $\task\test$.}
\label{tab:cp}
\centering
\scalebox{1}{
\begin{tabular}{c|cccc}
  & NN-IC & RW-IC &MTPS0 & MTPS+ \\
  \hline
  Cost  & 0.39 & 0.4 & 0.33 & \textbf{0.30}\\
\end{tabular}
}
\end{table}
Tab.~\ref{tab:cp} summarizes the expected costs across all test tasks
$\vec\task\test$. We averaged over all test tasks and 100 applications
of the learned policy, where the initial state was sampled from
$p(\vec x_0)$. 
%
Although NN-IC and RW-IC performed swing-up reliably, they incurred
the largest cost: For most test tasks, they balanced the pendulum at
the wrong cart position as they could not generalize from training
tasks to unseen test tasks.
In the MTPS experiments, the average cost was lowest, indicating that
our multi-task policy search approach is beneficial. MTPS+ led to the
best overall generalization performance, although it might not solve each
individual test task optimally.

\begin{figure}[tb]
\centering
\includegraphics[width = 0.9\hsize]{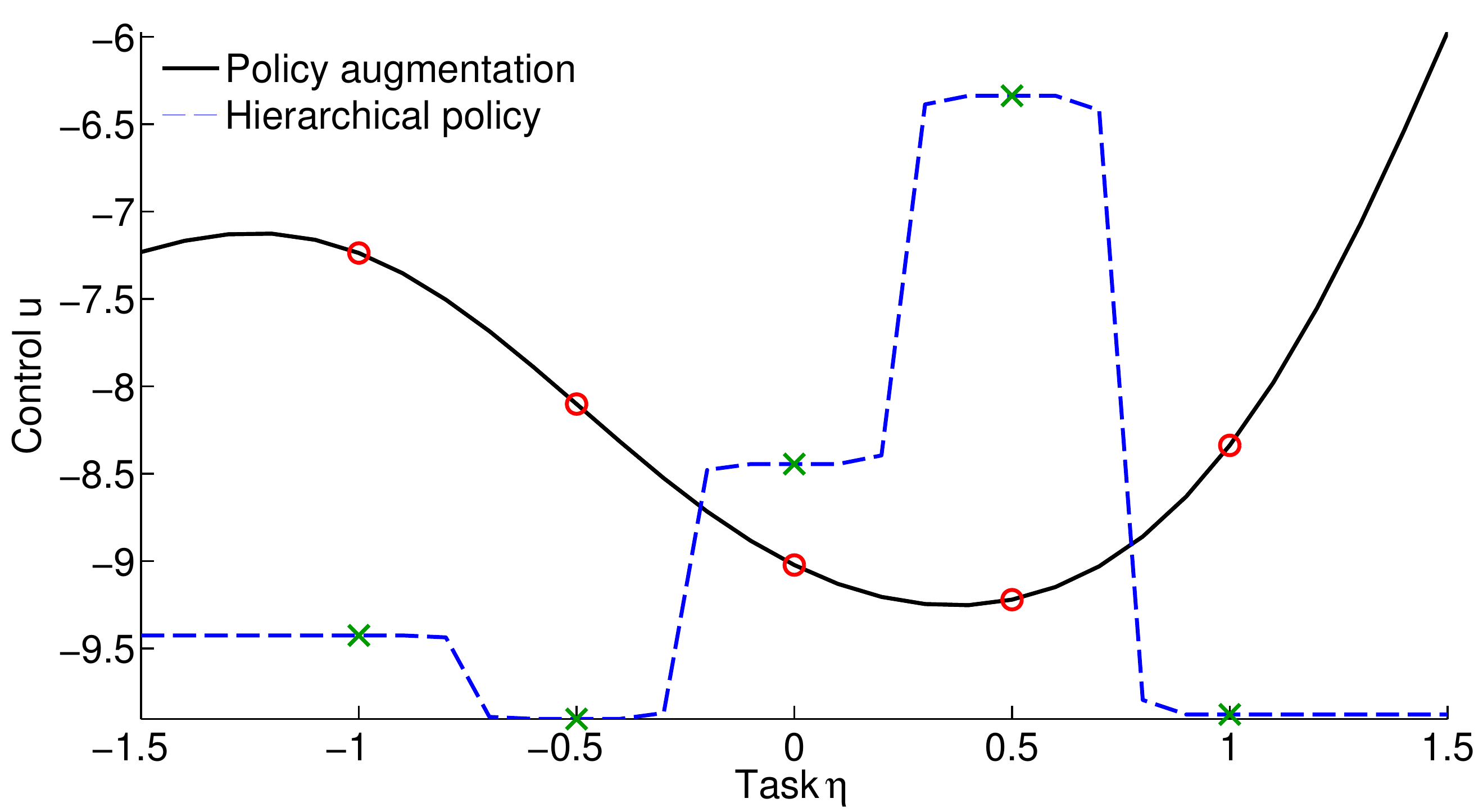}
\caption{Two multi-task policies for a given state $\vec x$, but
  varying task $\eta$. The black policy is obtained by applying our
  proposed multi-task approach (MTPS+); the blue, dashed policy is
  obtained by hierarchically combining local controllers (RW-IC). The
  training tasks are $\eta=\{\pm 1, \pm 0.5, 0\}$. The corresponding
  controls $u = \pi(\vec x,\eta)$ are marked by the red circles
  (MTPS+) and the green stars (RW-IC), respectively. MTPS+ generalizes
  more smoothly across tasks, whereas the hierarchical combination of
  independently trained local policies does not generalize well. }
\label{fig:multi-target motivation 2}
\end{figure}
Fig.~\ref{fig:multi-target motivation 2} illustrates the difference in
generalization performance between our MTPS+ approach and the RW-IC
approach, where controls $u_i$ from local policies $\pi_i$ are
combined by means of a gating network. Since the local policies are
trained independently, a (convex) combination of local controls makes
only sense in special cases, e.g., when the local policies are linear
in the parameters. In this example, however, the local policies are
nonlinear. Since the local policies are learned independently, their
overall generalization performance is poor. On the other hand, MTPS+
learns a single policy for a task $\eta_i$ always in the light of all
other tasks $\eta_{j\neq i}$ as well, and, therefore, leads to an
overall smooth generalization.

\subsection{Multi-Task Robot Manipulator}
%
\begin{figure}[tb]
\centering
\subfigure[Low-cost manipulator by Lynxmotion~\cite{lynxmotion}
  performing a block-stacking task. ]{
\includegraphics[width = 0.8\hsize]{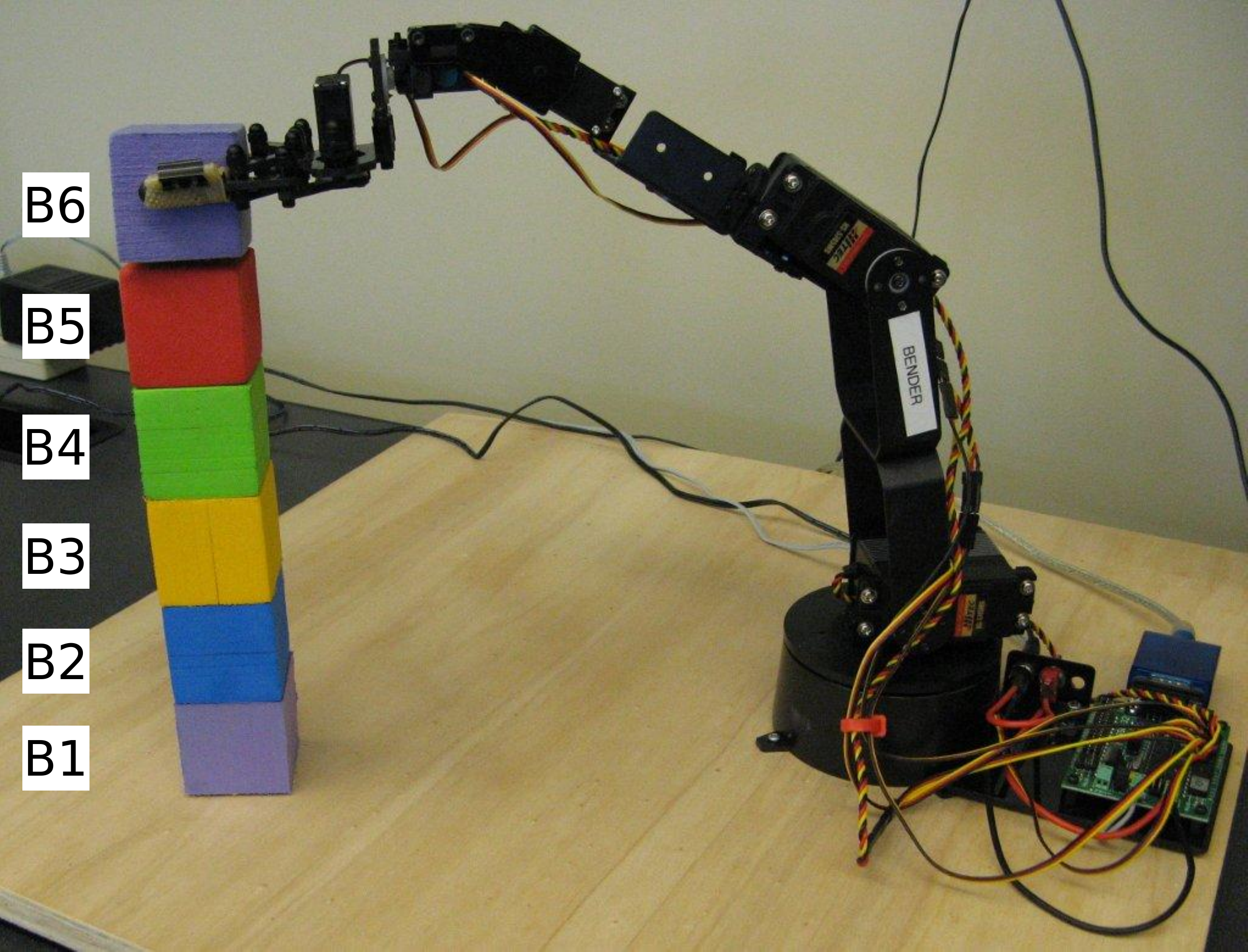}
\label{fig:robotic system}
}
\hfill
\subfigure[Average distances of the block in the gripper from the
    target position (with twice the standard error).]{
\includegraphics[width = 0.8\hsize]{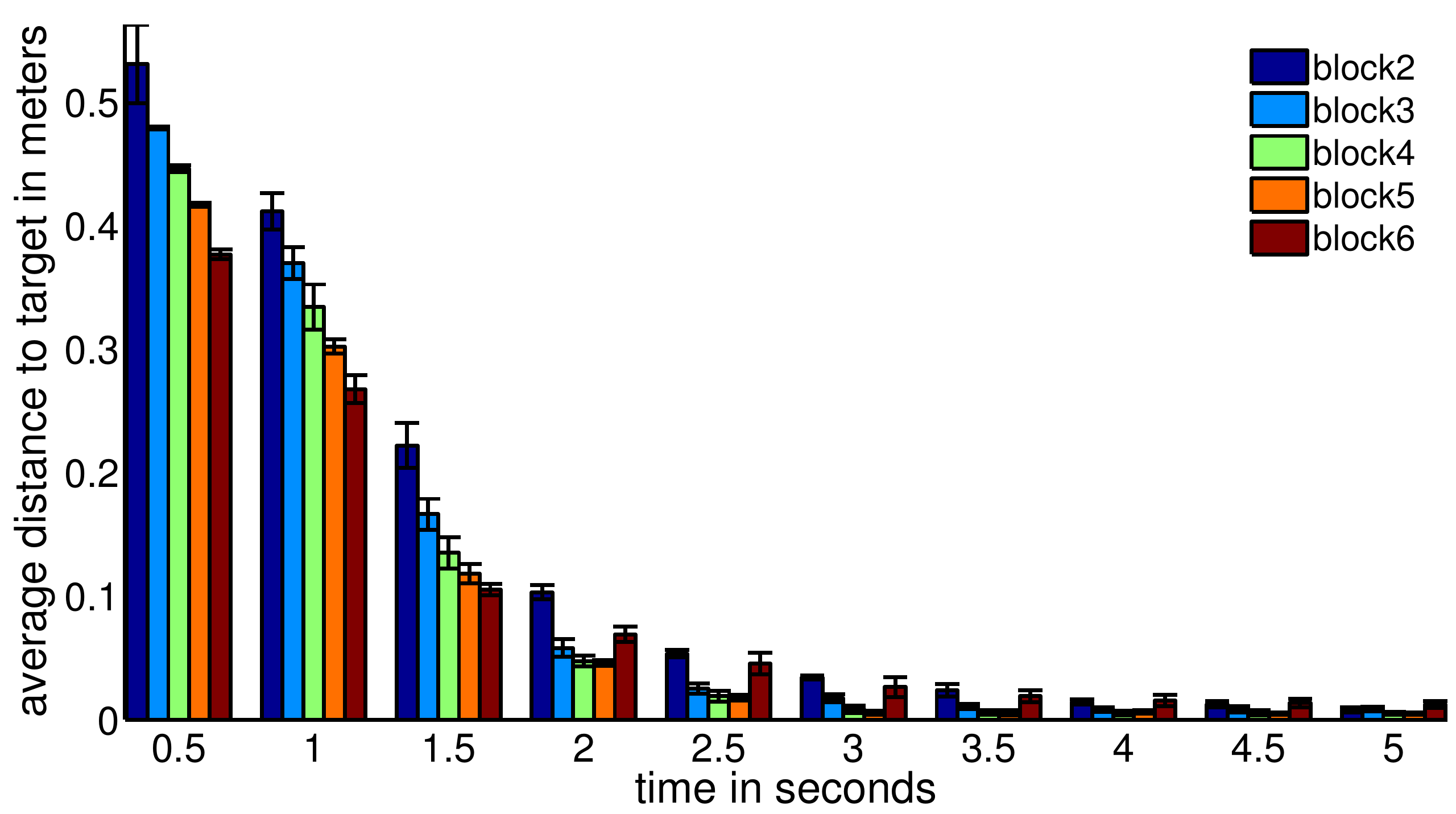}
\label{fig:distances block stacking}
}
\caption{Experimental setup and results for the multi-task
  block-stacking task. A controller was learned directly in the task
  space using visual feedback from a PrimeSense depth
  camera.}
\figspace
\end{figure}
Our proposed multi-task learning method has been applied to a
block-stacking task using a low-cost, off-the-shelf robotic
manipulator (\$370) by Lynxmotion~\cite{lynxmotion}, see
Fig.~\ref{fig:robotic system}, and a PrimeSense~\cite{primesense}
depth camera (\$130) used as a visual sensor.
%
The arm had six controllable degrees of freedom: base rotate, three
joints, wrist rotate, and a gripper (open\slash close). The plastic
arm could be controlled by commanding both a desired configuration of
the six servos (via their pulse durations) and the duration for
executing the command~\cite{Deisenroth2011b}.  
The camera was identical to the Kinect sensor, providing a
synchronized depth image and a $640\times 480$ RGB image at
$\unit[30]{Hz}$.  We used the camera for 3D-tracking of the block in
the robot's gripper.

The goal was to make the robot learn to stack a tower of six blocks
using multi-task learning.  The cost function $c$ in
Eq.~(\ref{eq:expected return}) penalized the distance of the block in
the gripper from the desired drop-off location.
We only specified the 3D camera coordinates of the blocks B2, B4, and
B5 as the training tasks $\vec\task\train$, see Fig.~\ref{fig:robotic
  system}. Thus, at test time, stacking B3 and B6 required exploiting
the generalization of the multi-task policy search. We chose $g(\vec
x,\vec\task) = \vec\task -\vec x$. Moreover, we set $\mat\Sigma^\task$
such that the task space, i.e., all 6 blocks, was well covered.  The
mean $\vec\mu_0$ of the initial distribution $\prob(\vec x_0)$
corresponded to an upright configuration of the arm.

A GP dynamics model was learned that mapped the 3D camera coordinates
of the block in the gripper and the commanded controls at time $t$ to
the corresponding 3D coordinates of the block in the gripper at time
$t+1$, where the control signals were changed at a rate of
$\unit[2]{Hz}$. Note that the learned model is not an inverse
kinematics model as the robot's joint state is unknown.
We used an affine policy $\vec u_t = \pi(\vec
x_t,\vec\target,\vec\polpar) = \mat A\vec x_t^{x,\target} + \vec b$,
where $\vec\polpar = \{\mat A, \vec b\}$.  The policy now defined a
mapping $\pi:\R^6\to\R^4$, where the four controlled degrees of
freedom were the base rotate and three joints.

We report results based on 16 training trials, each of length
$\unit[5]{s}$, which amounts to a total experience of $\unit[80]{s}$
only.
The test phase consisted of 10 trials per stacking task, where the arm
was supposed to stack the block on the currently topmost block. The
tasks $\vec\target\test_j$ at test time corresponded to stacking
blocks B2--B6 in Fig.~\ref{fig:robotic system}.
\begin{figure*}[tb]
\centering
\subfigure[Set-up for the imitation learning experiments. The orange
  balls represent the three training tasks $\vec\task_i\train$. The
  blue rectangle indicates the regions of the test tasks
  $\vec\task\test_j$ for our learned controller to which we want to
  generalize.]{
\includegraphics[height = 5.5cm]{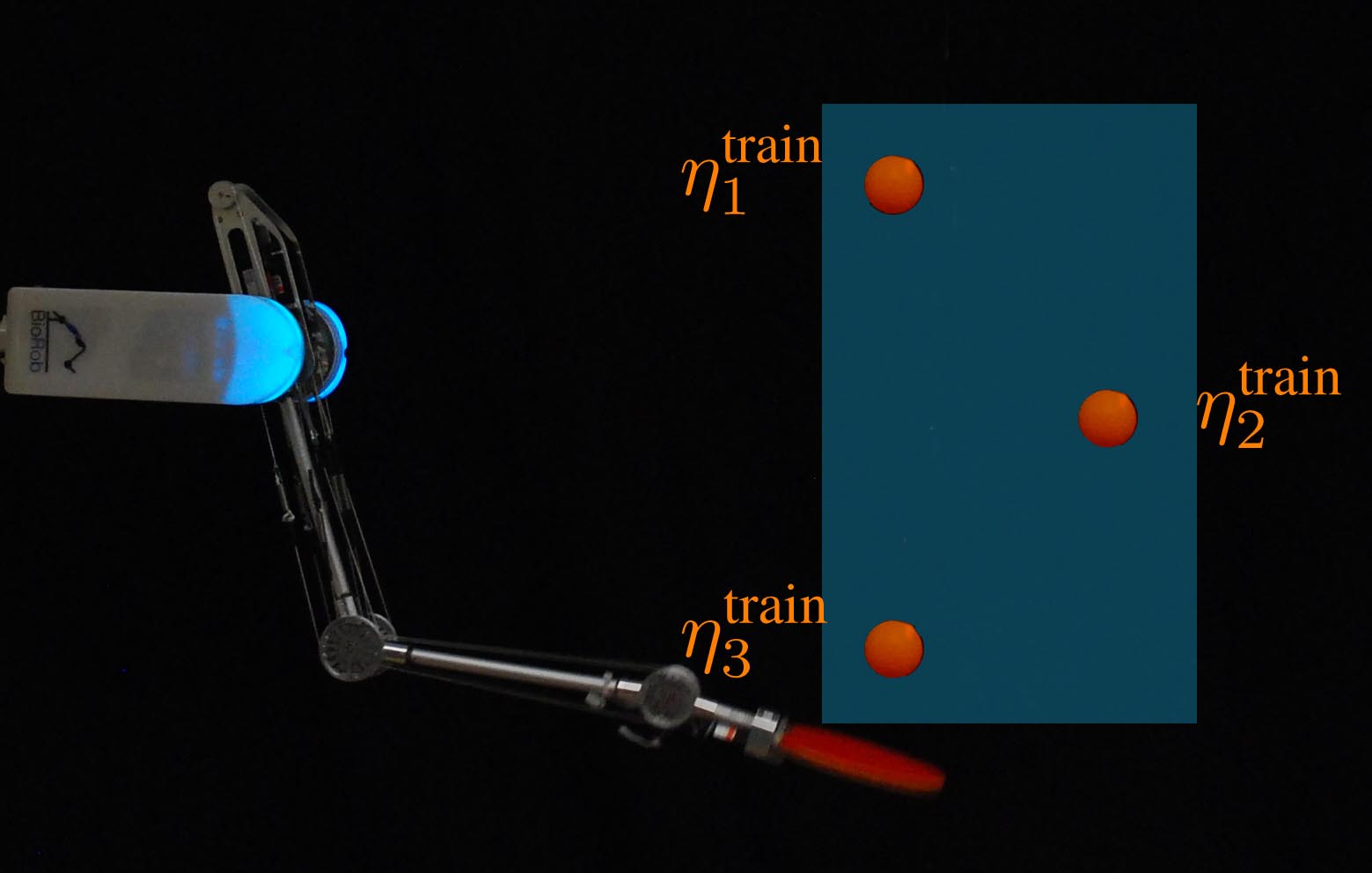}
\label{fig:biorob_setup}
}\hspace{8mm} \subfigure[The white discs are the training task
locations.  Blue and cyan indicate that the task was solved
successfully.]{
\includegraphics[height = 5.5cm]{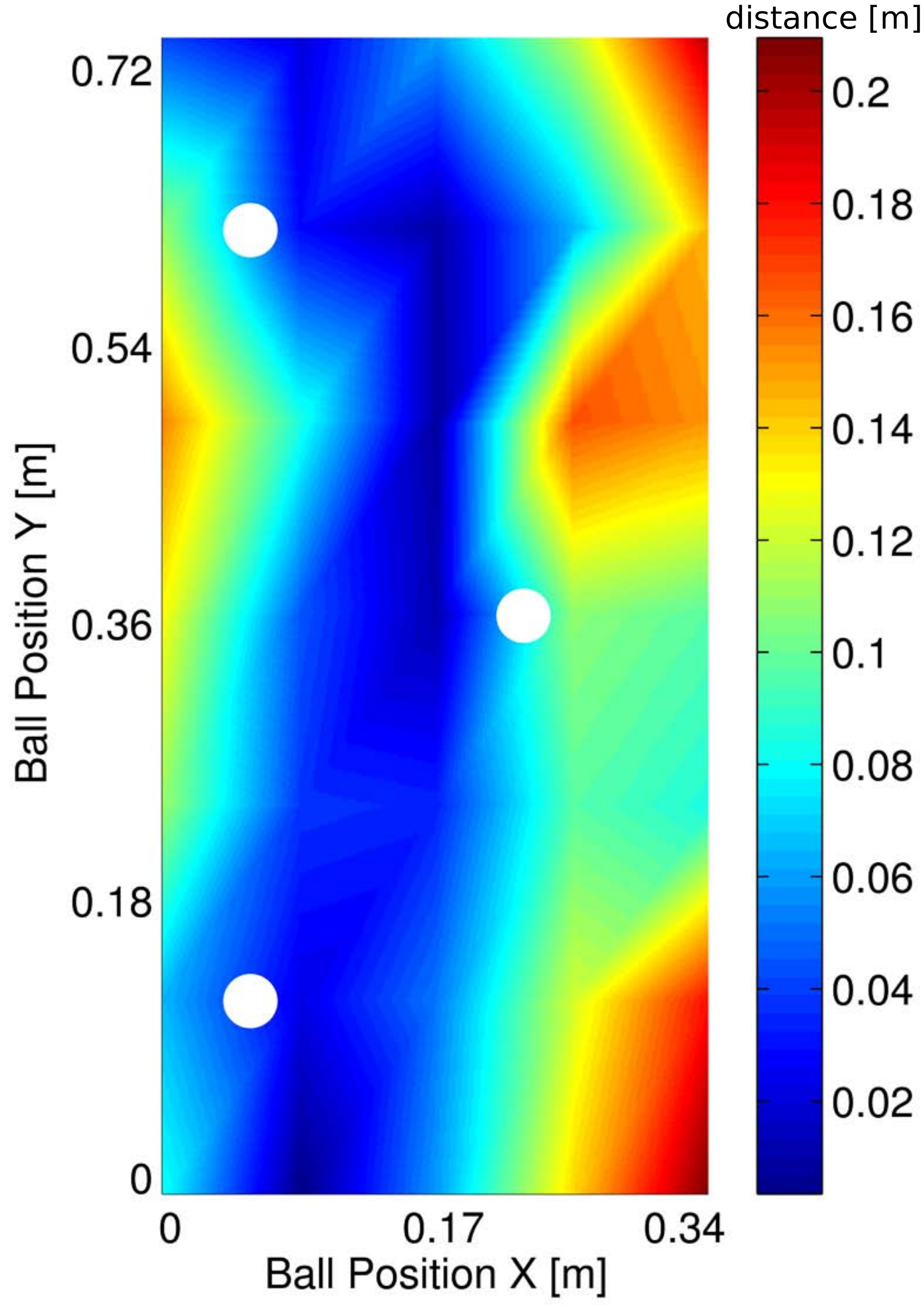}
\label{fig:biorob_results}
}
\vspace{-2mm}
\caption{Set-up and results for the imitation learning experiments
  with a bio-inspired BioRob\texttrademark.}
\figspace
\end{figure*}
Fig.~\ref{fig:distances block stacking} shows the average distance of
the block in the gripper from the target position, which was
$b=\unit[4.3]{cm}$ above the topmost block. Here, ``block2'' means
that the task was to move block B2 in the gripper on top of block
1. The horizontal axis shows times at which the manipulator's control
signal was changed (rate $\unit[2]{Hz}$), the vertical axis shows the
average distances (over 10 test trials) to the target position in
meters. For all blocks (including blocks B3 and B6, which were not
part of the training tasks $\vec\task\train$) the distances approached
zero over time. Thus, the learned multi-task controller was able to
interpolate (block B3) and extrapolate (block B6) from the training
tasks to the test tasks \emph{without} re-training.

\subsection{Multi-Task Imitation Learning of Ball-Hitting Movements}
\label{sec:imitation learning}

We demonstrate that our MTPS approach can also be applied to imitation
learning.  Instead of defining a cost function $c$ in
\eq~(\ref{eq:expected return}), a teacher provides demonstrations that
the robot should imitate. We show that our MTPS approach allows to
generalize from demonstrated behavior to behaviors that have not been
observed before.
In~\cite{Englert2013}, we developed a method for model-based imitation
learning based on probabilistic trajectory matching for a single
task. The key idea is to match a distribution over predicted robot
trajectories $p(\vec \tau\pred)$ directly with an observed distribution
$p(\vec\tau\expert)$ over expert trajectories $\vec
\tau\expert$ by finding a policy $\pi^*$ that
minimizes the KL divergence \cite{kullback1959information} between
them. 

In this paper, we extend this imitation learning approach to a
multi-task scenario to jointly learning to imitate multiple tasks from
a small set of demonstrations. In particular, we applied our
multi-task learning approach to learning a controller for hitting
movements with variable ball positions in a 2D-plane using the
tendon-driven BioRob\texttrademark~X4, a five DoF compliant,
light-weight robotic arm, capable of achieving high accelerations, see
Fig.~\ref{fig:biorob_setup}.
The torques are transferred from the motor to the joints via a system
of pulleys, drive cables, and springs, which, in the
biomechanically-inspired context, represent tendons and their
elasticity.  While the BioRob's design has advantages over traditional
approaches, modeling and controlling such a compliant system is
challenging.

In our imitation-learning experiment, we considered three joints of
the robot, such that the state $\vec x \in \R^6$ contained the joint
positions $\vec q$ and velocities $\dot{\vec q}$ of the robot. The
controls $\vec u \in \R^3$ were given by the corresponding motor
torques, directly determined by the policy $\pi$.  For learning a
controller, we used an RBF network with $250$ Gaussian basis
functions, where the policy parameters comprised the locations of the
basis functions, their weights, and a shared diagonal covariance
matrix, resulting in about $2300$ policy parameters. Policy learning
required about 20 minutes computation time.
Unlike in the previous examples, we represented a task as a
two-dimensional vector $\vec \task \in \R^2$ corresponding to the ball
position in Cartesian coordinates in an arbitrary reference frame
within the hitting plane. As the task representation $\vec\task$ was
basically an index and, hence, unrelated to the state of the robot,
$g(\vec x,\vec\task) = \vec\task$, and the cross-covariances $\mat
C^{x\task}$ in \eq~(\ref{eq:augmented state distribution}) were $\mat
0$.

As training tasks $\vec\task_j\train$, we defined hitting movements
for three different ball positions, see
\fig~\ref{fig:biorob_setup}. For each training task, an expert
demonstrated two hitting movements via kinesthetic teaching. Our goal
was to learn a single policy that a) learns to imitate three distinct
expert demonstrations, and b) generalizes from demonstrated behaviors
to tasks that were not demonstrated. In particular, these tests tasks
were defined as hitting balls in a larger region around the training
locations, indicated by the blue box in
\fig~\ref{fig:biorob_setup}. We set the matrices $\mat\Sigma^\task$
such that the blue box was covered well.

\fig~\ref{fig:biorob_results} shows the performance results as a
heatmap after $15$ iterations of \alg~\ref{alg:pilco}. The evaluation
measure was the distance in $\unit{m}$ between the ball position and
the center of the table-tennis racket. We computed this error in a
regular 7x5 grid of the blue area in \fig~\ref{fig:biorob_setup}. The
distances in the blue and cyan areas were sufficient to successfully
hit the ball (the racket's radius is about $\unit[0.08]{m}$). Hence,
our approach successfully generalized from given demonstrations to new
tasks that were not in the library of demonstrations.

\subsection{Remarks}

Controlling the cart-pole system to different target location is a
task that could be solved without the task-augmentation of the
controller inputs: It is possible to learn a controller that depends
only on the position of the cart relative to the target location---the
control signals should be identical when the cart is at location
$\chi$, the target location is at $\chi+\varepsilon$ or when the cart is
at location at position $x_2$, and the target location is at
$x_2+\varepsilon$. Our approach, however, learns these invariances
automatically, i.e., it does not require an intricate knowledge of the
system\slash controller properties. Note that a linear combination of
local controllers usually does not lead to success in the cart-pole
system, which requires a nonlinear controller for the joint task of
swinging up the pendulum and balancing it in the inverted position.

In the case of the Lynx-arm, these invariances no longer exist as the
optimal control signal depends on the absolute position of the arm,
not only on the relative distance to the target. Since a linear
controller is sufficient to learn block stacking, a convex combination
of individual controllers should be able to generalize from the
trained blocks to new targets if no extrapolation is required.


\section{Conclusion and Future Work}


We have presented a policy-search approach to multi-task learning for
robots, where we assume stationary dynamics. Instead of combining
local policies using a gating network, which only works for
linear-in-the-parameters policies, our approach learns a single policy
jointly for all tasks. The key idea is to explicitly parametrize the
policy by the task and, therefore, enable the policy to generalize
from training tasks to similar, but unknown, tasks at test time. This
generalization is phrased as an optimization problem, jointly with
learning the policy parameters. For solving this optimization problem,
we incorporated our approach into the \textsc{pilco} policy search
framework, which allows for data-efficient policy learning. We have
reported promising results on multi-task RL on a standard benchmark
problem and on a robotic manipulator. Our approach also applies to
imitation learning and generalizes imitated behavior to solving tasks
that were not in the library of demonstrations.


%
In this paper, we considered the case that re-training the policy
after a test run is not allowed. Relaxing this constraint and
incorporating the experience from the test trials into a subsequent
iteration of the learning procedure would improve the average quality
of the controller.

In future, we will jointly learn the task representation for the
policy and the policy parametrization. Thereby, it will not be
necessary to specify any interdependence between task and state space
a priori, but this interdependence will be learned from data.



\end{document}